\documentclass[conference]{IEEEtran}
\IEEEoverridecommandlockouts
\usepackage{cite}
\usepackage{amsmath,amssymb,amsfonts}
\usepackage{algorithmic}
\usepackage{graphicx}
\usepackage{textcomp}
\usepackage{multirow}
\usepackage{xcolor}

\def\BibTeX{{\rm B\kern-.05em{\sc i\kern-.025em b}\kern-.08em
    T\kern-.1667em\lower.7ex\hbox{E}\kern-.125emX}}
\begin{document}

\title{Federated Anomaly Detection over Distributed Data Streams    \\
\thanks{This work is aligned with project AIDA (Adaptive, Intelligent, and Distributed Assurance Platform) project (reference code: 45907)}
}

\author{\IEEEauthorblockN{Paula Raissa Silva}
\IEEEauthorblockA{\textit{FEUP, INESC TEC} \\
Porto, Portugal \\
paula.r.silva@inesctec.pt}
\and
\IEEEauthorblockN{João Vinagre}
\IEEEauthorblockA{\textit{FCUP, INESC TEC} \\
Porto, Portugal \\
joao.m.silva@inesctec.pt}
\and
\IEEEauthorblockN{João Gama}
\IEEEauthorblockA{\textit{FEP, INESC TEC} \\
Porto, Portugal \\
jgama@fep.up.pt}
}

\maketitle

\begin{abstract}
Sharing of telecommunication network data, for example, even at high aggregation levels, is nowadays highly restricted due to privacy legislation and regulations and other important ethical concerns. 
It leads to scattering data across institutions, regions, and states, inhibiting the usage of AI methods that could otherwise take advantage of data at scale.
It creates the need to build a platform to control such data, build models or perform calculations.
In this work, we propose an approach to building the bridge among anomaly detection, federated learning, and data streams.  
The overarching goal of the work is to detect anomalies in a federated environment over distributed data streams.  
This work complements the state-of-the-art by adapting the data stream algorithms in a federated learning setting for anomaly detection and by delivering a robust framework and demonstrating the practical feasibility in a real-world distributed deployment scenario.
\end{abstract}


\section{Problem Definition and Motivation}

The identification of anomalies in data streams collected and transformed by distributed data points has attracted significant research interest.
Traditional anomaly detection techniques were built to deal with static data and require a complete dataset for building models.
These techniques may not be adequate to work over data streams because the entire stream can not be stored, and the stream data distribution may change over time.
Then, there is the need to develop strategies for anomaly detection over data streams. 

With the expansion of digitization, the emergence of heterogeneous and intelligent devices in everyday life has led to a massive increase in available data assets.
The purpose of the privacy-sensitive analysis is to use this data to its full potential without compromising privacy.
Data analysis that preserves privacy has become a crucial aspect and is recognized as a significant problem in the data processing.

An emerging branch within AI, and Machine Learning, is Federated Learning (FL) \cite{DBLP:journals/kbs/ZhangXBYLG21}.
It is a decentralized approach in which the learning of models is pushed towards the network nodes that collect and hold the data. 
This is an inherently incremental and distributed process, and therefore it is well suited to learn from distributed data streams and static data in remote repositories. 
Federated platforms may deal with reducing data transfer, lowering latency, reducing storage and network bandwidth requirements, improving the overall quality of service while at the same time providing an additional layer of privacy since data never leaves the origin \cite{yang2019federated}.

The overarching goal of the proposal is to detect anomalies in a federated environment over distributed data streams.
As an application area, we choose telecommunications networks due to their distributed architecture, large flow, and privacy concerns.
The platform may enable model training, testing, and evaluation in a federated environment over distributed data streams.

There are several challenges associated with federated anomaly detection over distributed data streams.
Such problems could not be entirely solved with state-of-the-art techniques, and this proposal addresses these new challenges from the followings research questions (RQ): 

\begin{itemize}
    \item[RQ1] How do we find anomalous behaviors in distributed data nodes from a telecommunications network without transferring individual-level data among participants? To answer this question, we propose the study and building a framework to integrate and make available privacy-preserving methodologies and algorithms to detect anomalous behaviors over distributed data streams.
    \item[RQ2] Could the techniques and algorithms for data streams be adapted to a federated environment? What is the ideal outline to train federated models over data streams when computational resources are limited? Most of the algorithms for data streams consider the learning process in a centralized environment, and even the distributed streaming algorithms do not consider both privacy-preserving and computational constraints. We propose studying the reformulation of existing algorithms for online anomaly detection and federated anomaly detection to address this question.
    \item[RQ3]Are traditional evaluation protocols suitable for anomaly detection over dynamic federated environments? If not, how do we evaluate anomaly detection algorithms in such environments? In order to evaluate algorithms, we need an evaluation methodology that enables a low resource consumption and accurate comparison among algorithms for anomaly detection over federated data streams.
\end{itemize}

In the end, the expected contributions are: a framework with a set of federated algorithms for anomaly detection over distributed data streams; a federated architectural outline containing the framework and additional components that can be easily deployed; an evaluation methodology specifically for data streams algorithms in federated environments.

\section{Proposal Outline}

This proposal aims to develop federated learning (FL) methods by adapting existing machine learning algorithms or developing new ones to work on an FL framework.
The proposed approach contemplates the following activities: framework development, algorithms for federated anomaly detection over data streams, and evaluation methods.
As an application area, we consider telecommunications use cases, such as service disruption detection and service abuse, which should be attended in the experimental step to validate the proposed approach.
Finally, the research product will be a system prototype able to detect anomalies over data streams in a federated environment.

\subsection{Framework Development}
In this step, we will propose the modules of a framework for integrating and make available trustworthy methodologies and algorithms for anomaly detection.
The approach should deal with real-time data in the context of federated learning.
The framework may provide an integrated view of the learning process.

For this step, the proposal involves the following tasks: (i) define the federated architecture; (ii) privacy of federated learning; (iii) data integration; (iv) Interfaces for model developers and model clients; (v) Access the framework in terms of energetic efficiency, usability, and interoperability.
The main outcome of this task will be the modules of a framework for integrating and make available trustworthy methodologies and algorithms for federated anomaly detection over data streams.

\subsection{Algorithms for Federated Anomaly Detection over Data Streams}

The study of algorithms for federated anomaly detection over data streams step aims to build models capable of identifying anomalies in telecommunications network data. 
This activity will study the reformulation of existing algorithms for online anomaly detection and federated anomaly detection.
We also consider the development of new techniques for federated online learning anomaly detection.
The goal is to build the bridge between online learning and federated learning.
The central idea is based on the computation of aggregated values on each data node, and the global model will be computed at the central node to minimize the number of calls to each data node.

For this step, the proposal involves the following tasks: (i) stream-based algorithms definition and implementation in the federated framework; (ii) distributed learning with computational and energy constraints; (iii) compliance with privacy regulations and best practices.
The main outcome of this task will be a set of algorithms able to learn local models at the edge that can be combined in a federated learning scheme and with the ability to run under privacy and computational constraints.

\subsection{Evaluation Methods}

The goal of evaluating the federated learning model is to find the ideal metrics for each type of problem.
This activity will focus on the study of evaluation techniques for data streams and federated learning.
For this step, the proposal involves the following tasks: (i) statistical metrics; (ii) system metrics.
The expected outcome is a set of system and statistical metrics which best evaluate and validate models built with online anomaly detection algorithms in a federated environment.

\section{Relevant Related Work and Novelty}

\begin{table}[h!]
\begin{tabular}{|l|c|p{4.3cm}|}
\hline & Reference & Method \\ \hline
\multicolumn{1}{|c|}{\multirow{3}{*}{\begin{tabular}[c]{@{}c@{}}Stream-based \\ aproaches\end{tabular}}} &  \cite{Rettig2015}      &Framework for anomaly detection over high-velocity streams of mobile network data.       \\ \cline{2-3} 
\multicolumn{1}{|c|}{} &    \cite{veloso2020case}       &  Hierarchical heavy hitter algorithm forfraud detection in phonecalls.  \\ \cline{2-3} 
\multicolumn{1}{|c|}{} &    \cite{Liang2021}       & Budget online learning algorithm to detect anomalies in imbalanced data streams \\ \hline
\multirow{3}{*}{\begin{tabular}[c]{@{}c@{}}Federated learning\\ for anomaly \\ detection\end{tabular}}   &    \cite{Preuveneers2018}       &    Deep neural network for anomalydetection in a blockchain    \\ \cline{2-3} 
  &    \cite{Nguyen2019}       & Anomalous self-learning for malware detection in IoT devices.  \\ \cline{2-3} 
&    \cite{Zhao2019}  &    Multi-task deep neural network (MT-DNN-FL) federated network traffic.    \\ \cline{2-3} 
& \cite{hao2019efficient} & Convolutional neural  network-longshort-term memory (AMCNN-LSTM) forIndustrial IoT Networks.   \\
\hline
\end{tabular}
\end{table}

\bibliographystyle{IEEEtran}
\bibliography{myrefs}

\end{document}